\documentclass[10pt,twocolumn,letterpaper]{article}

\usepackage{iccv}
\usepackage{times}
\usepackage{epsfig}
\usepackage{graphicx}
\usepackage{amsmath}
\usepackage{amssymb}
\usepackage{booktabs}

\usepackage[pagebackref=true,breaklinks=true,letterpaper=true,colorlinks,bookmarks=false]{hyperref}

\iccvfinalcopy 

\def\iccvPaperID{3971} 

\ificcvfinal\fi

\usepackage{amsthm}
\newtheorem{Def}{Definition}
\newtheorem{Cor}{Corollary}
\newtheorem{Them}{Theorem}
\newtheorem{Lemma}{Lemma}

\begin{document}

\title{Breadcrumbs: Adversarial Class-Balanced Sampling for Long-tailed Recognition}

\author{Bo Liu\\
UC, San Diego\\
{\tt\small boliu@ucsd.edu}
\and
Haoxiang Li\\
Wormpex AI Research\\
{\tt\small lhxustcer@gmail.com}
\and
Hao Kang\\
Wormpex AI Research\\
{\tt\small haokheseri@gmail.com}
\and
Gang Hua\\
Wormpex AI Research\\
{\tt\small ganghua@gmail.com}
\and
Nuno Vasconcelos\\
UC, San Diego\\
{\tt\small nuno@ece.ucsd.edu}
}

\maketitle
\ificcvfinal\thispagestyle{empty}\fi

\begin{abstract}
   The problem of long-tailed recognition, where the number of examples per class is highly unbalanced, is considered. While training with class-balanced sampling has been shown effective for this problem, it is known to over-fit to few-shot classes.  It is hypothesized that this is due to the repeated sampling of examples and can be addressed by feature space augmentation. A new feature augmentation strategy, EMANATE, based on back-tracking of features across epochs during training, is proposed. It is shown that, unlike class-balanced sampling, this is an adversarial augmentation strategy. A new sampling procedure, Breadcrumb, is then introduced to implement adversarial class-balanced sampling without extra computation. 
   Experiments on three popular long-tailed recognition datasets show that Breadcrumb training produces classifiers that outperform existing solutions to the problem.
\end{abstract}


\section{Introduction}

The availability of large-scale datasets, with many images per class~\cite{imagenet_cvpr09}, has been a major factor in the success of deep learning for computer vision. However, these datasets are manually curated and artificially balanced. 
This is unlike most real world applications, where the frequencies of examples from different classes can be highly unbalanced, leading to skewed distributions with long tails. These datasets are composed by a few popular classes and many rare classes.
This class imbalance has been observed in image classification~\cite{wang2016learning}, face identification~\cite{huang2019deep,liu2015deep}, object detection~\cite{lin2017focal,zhu2014capturing}, and many other applications. Researchers have tackled it from various angles, including zero-shot learning~\cite{felix2018multi,xian2018feature,xian2019f}, few-shot learning~\cite{vinyals2016matching,snell2017prototypical,finn2017model}, and more recently long-tailed recognition~\cite{liu2019large}. 

\begin{figure}[t!]
\centering{
	\includegraphics[width=0.4\textwidth]{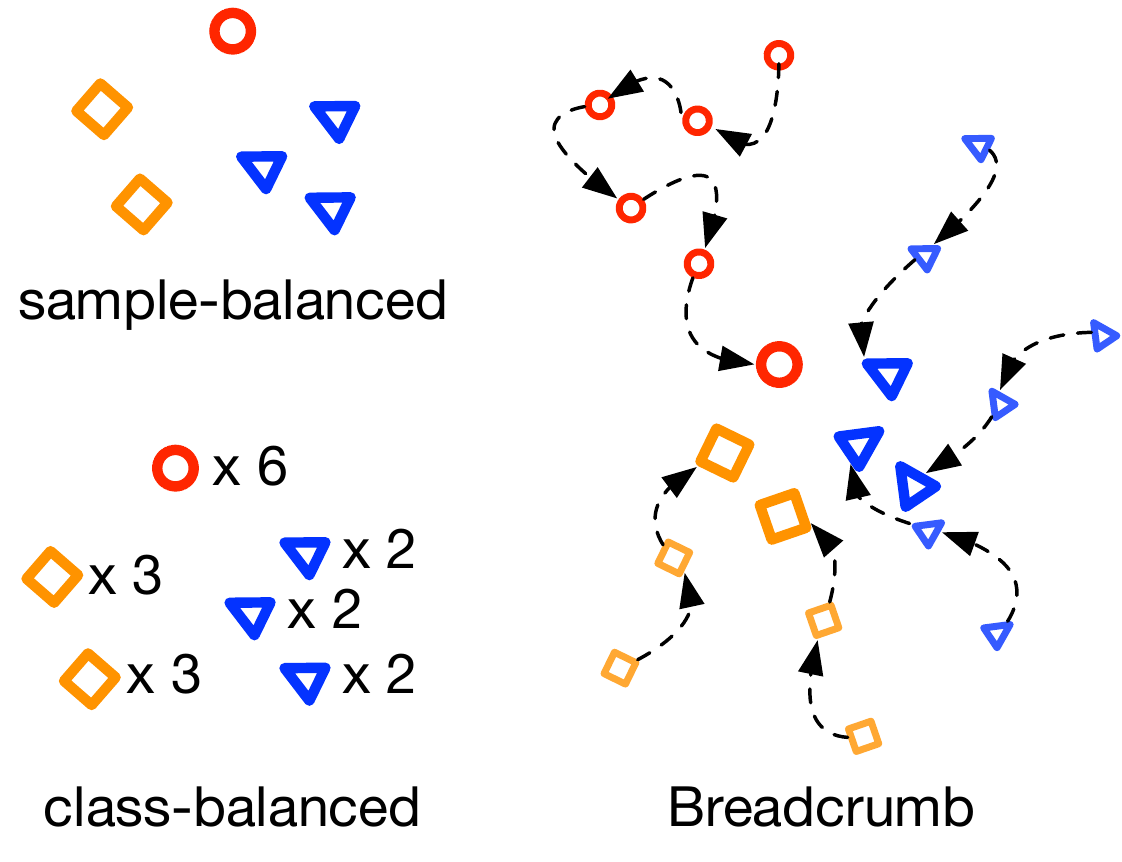}
	}
	\caption{Upper-Left: Random sampling is sample-balanced. The number of examples per class has a long-tailed distribution. This leads to under-fitting in few-shot classes. Lower-Left: Class-balanced sampling duplicates few-shot samples in feature space and can leads to over-fitting for these classes. Right: Breadcrumb produces trails of features by back-tracking through training epochs. This is shown to be an adversarial augmentation technique, which mitigates the over-fitting problem.}
	\label{fig:breadcrumb}
\end{figure}

In this work, we focus on the long-tailed recognition setting, where classes are grouped into three types that differ in training sample cardinality: many-shot ($>100$ samples), medium-shot (between $20$ and $100$ samples), and few-shot ($\leq20$ samples). Performance is evaluated over each group independently, in addition to the overall classification accuracy. While training data is highly unbalanced, the test set is kept balanced so that equally good performance on all classes is a requisite for high accuracy. 
One of the insights from the long-tailed recognition literature is that techniques targeting specific dataset limitations, e.g. few-shot learning by data augmentation~\cite{chen2019image,wang2018low}, predicting classifier weights~\cite{qi2018low}, prototype-based non-parametric classifiers~\cite{snell2017prototypical}, and optimization with second derivatives~\cite{finn2017model}, are frequently harmful to classes that do not suffer from those limitations, e.g. many-shot. Hence, it is important to address the problem holistically, considering all types of classes simultaneously. 

Since long-tailed recognition datasets have a continuous coverage of the number of samples per class, they are best addressed by  training a  model on the entire dataset, in a way robust to data imbalance. Standard classifier training follows the {\it sample-balanced\/} sampling setting of Figure~\ref{fig:breadcrumb}. This consists of sampling images uniformly to create batches for network training. In result, as shown in the figure, few-shot classes (red) are under-represented and many-shot classes (blue) are over-represented in each batch. Hence, learning typically {\it under-fits} less populated classes. This has motivated procedures to fight class imbalance with data re-sampling~\cite{zhang2017range} or cost-sensitive losses~\cite{lin2017focal} that place more training emphasis on examples of lower populated classes. 
One of the more successful approaches is to decouple the training of feature embedding and classifier~\cite{kang2019decoupling}. While the embedding is learned with image-balanced training, the classifier is trained with {\it class-balanced\/} sampling. As illustrated in Figure~\ref{fig:breadcrumb}, this consists of sampling classes uniformly and then sampling uniformly within the class. However, for few shot classes, this approach leads to repeated sampling of the same examples. In result, the classifier can easily {\it over-fit} on few-shot classes. 

In this work, we adopt the decoupled training strategy but seek to avoid over-fitting in the classifier training stage. For this, we propose to enrich the training data in the feature space at the output of the embedding, without extra computation. 
The idea is to back-track features to access the large diversity of feature vectors that are available per training image in prior epochs. This can be exploited to generate more diverse training data than simply replicating existing features. We refer to this procedure as {\it feature back-tracking\/}.  As shown in Figure~\ref{fig:breadcrumb}, it allows the sampling of large numbers of feature vectors from the few-shot classes without duplication. Since the embedding changes across training epochs, an alignment is necessary to simplify network training. We show that a simple alignment of class means suffices to accomplish this goal and propose the {\it fEature augMentAtioN by bAck-tracking wiTh alignmEnt\/} (EMANATE) procedure. This consists of augmenting the feature vectors collected at an epoch with aligned replicas of the vectors that emanate from them by back-tracking. 

A theoretical analysis shows that, unlike class-balanced sampling, EMANATE is an adversarial feature augmentation technique, in the sense that it is guaranteed to increase the training loss for any convergent training scheme. This places EMANATE in the realm of feature augmentation methods popular in the few-shot literature~\cite{chen2019image,wang2018low}. However, these require extra computation to generate new examples and sometimes introduce convergence problems.
EMANATE requires no extra computation and can be applied differently to each class, according to its number of samples. For classes with enough samples, only features from the last epoch are used, i.e. no resampling is performed. For those without, features are back-tracked over previous epochs, until there are enough features. This results in a new training feature set of higher variance for few-shot classes but forces no change on many-shot classes. In result, it is possible to improve classification accuracy for the former without degrading performance for the latter. 

A new sampling scheme, denoted {\it Breadcrumb Sampling\/} is then proposed to leverage the feature trails extracted by EMANATE, in the context of the two-stage training of class-balanced sampling. Breadcrumb Sampling relies on EMANATE to collect these feature trails in a first stage, when the embedding is trained with image-balanced sampling. In the second stage, the classifier is then learned with class-balanced training based on these trails. Two sampling variants are considered. Weak Breadcrumb Sampling only uses feature trails collected at the end of stage 1, i.e. once the embedding has converged. Strong Breadcrumb Sampling uses trails collected throughout  stage 1 training, i.e. as the embedding evolves. This tends to create an even more adversarial training set


\begin{figure*}[t!]
\centering{
	\includegraphics[width=\textwidth]{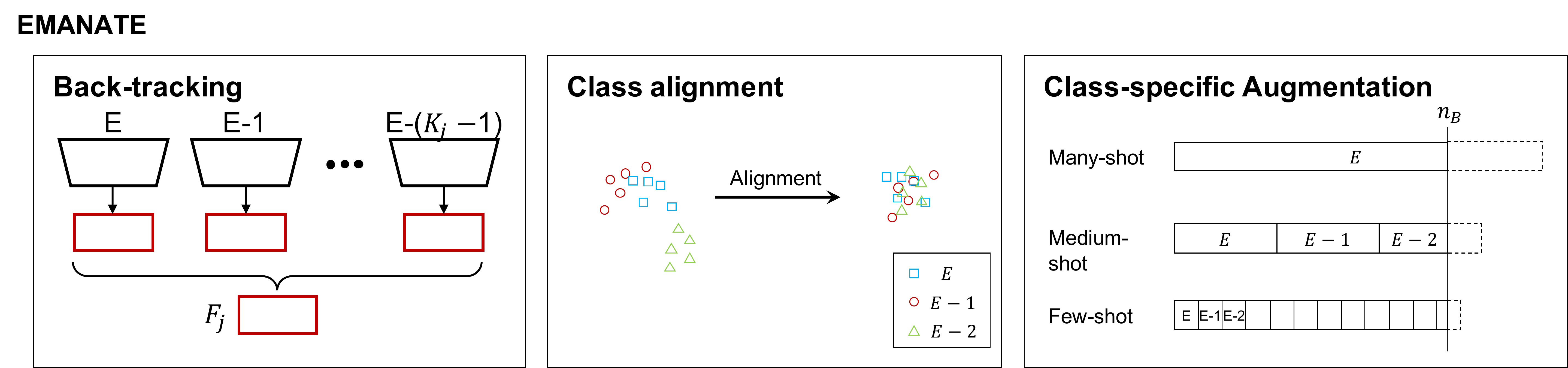}
	}
	\caption{EMANATE. Left: features from embeddings learned in previous epochs are back-tracked to compose a class-balance training set. Middle: Class alignment aligns the means of features from different epochs. Right: different classes have different back-tracking lengths. Many-shot classes only collect features from the current epoch; medium-shot classes back-track for a few epochs; and few-shot classes from many. When the number of samples exceeds $n_B$, the earliest epoch is randomly sampled to meet this target.}
	\label{fig:augmentation_and_hard}
\end{figure*}



Overall, this work makes several contributions. First, we point out that class-balanced sampling is not an adversarial augmentation technique, which limits its ability to combat over-fitting in few-shot classes.
Second, we propose EMANATE, a data augmentation technique that addresses this problem by feature back-tracking with alignment. Third, we show theoretically that, unlike class-balanced sampling, EMANATE is an adversarial technique. Fourth, we propose two variants of a new sampling scheme, Breadcrumbs, which leverage EMANATE to enable long-tailed recognition with state of the art performance.  
All of this is achieved with no extra computation and no performance degradation for classes with many examples.

\section{Related Work}

\noindent
{\bf Long-tailed recognition} 
has recently received substantial attention~\cite{wang2017learning,oh2016deep,lin2017focal,zhang2017range,liu2019large,wang2016learning}. Several approaches have been proposed, including metric learning~\cite{oh2016deep,zhang2017range}, loss weighting~\cite{lin2017focal}, or meta-learning~\cite{wang2016learning}. Some methods propose dedicated loss functions to mitigate the data imbalanced problem. For example, lift loss~\cite{oh2016deep} introduces margins between many training samples. Range loss~\cite{zhang2017range} encourages data from the same class to be close and different classes to be far away in the embedding space. The focal loss~\cite{lin2017focal} dynamically balances weights of positive, hard negative, and easy negative samples. As reported by Liu et al~\cite{liu2019large}, when applied to long-tailed recognition, many of these methods 
improved accuracy of the few-shot group, but at the cost of lower accuracy for many-shot classes. 

Other methods, e.g. class-balanced experts~\cite{sharma2020long} and knowledge distill~\cite{xiang2020learning}, try to mitigate this problem by artificially dividing the training data into subsets, based on number of examples, and training an expert per subset. However, experts learned from arbitrary data divisions can be sub-optimal, especially for few-shot classes, where training data is insufficient to learn the expert model. 

More recent works~\cite{kang2019decoupling,zhou2020bbn} achieve improved long-tailed recognition by training feature embedding and classifier with separate sampling strategies. The proposed Breadcrumbs approach follows this strategy, learning the embedding in a first stage with sample-balanced (random) sampling and the classifier in a second stage with class-balanced sampling. In fact, Breadcrumbs can be seen as a data augmentation method tailored for this strategy, improving its long-tailed recognition performance over all class groups.

Another related work is LEAP~\cite{liu2020deep}, a method mostly tested on person re-identification and face recognition problems, where datasets usually have long-tailed distributions. LEAP augments data samples from tail (few-shot) classes by transferring intra-class variations from head (many-shot) classes. This assumes a shared intra-class variation across classes, which can hold for person re-ID and face recognition but may not be applicable for general long-tailed recognition tasks. Besides, LEAP is technically orthogonal to Breadcrumbs and the two methods could potentially be combined for further improvement.

\noindent
{\bf Few-shot learning} focus solely on the data scarcity problem. A large group of approaches is based on meta-learning, using gradient based methods such as MAML and its variants~\cite{finn2017model,finn2018probabilistic}, or LEO~\cite{rusu2018meta}. These methods take advantage of second derivatives to optimize the model from few-shot samples. Another group of methods, including matching network~\cite{vinyals2016matching}, prototypical network~\cite{snell2017prototypical}, and relation network~\cite{sung2018learning},aims to learn robust metrics. Since these methods are designed specifically for few-shot classes, they often under-perform for many-shot classes, which makes them ineffective for long-tailed recognition.

Similarly to Breadcrumbs, some few-shot methods have proposed to augmenting training data by combining GANs with meta-learning~\cite{wang2018low}, synthesizing features across object views~\cite{Liu_2018_CVPR} or using other forms of data hallucination~\cite{hariharan2017low}. All these method introduces non-negligible extra computation to generate the new data samples. The application of GAN-based methods to few-shot data without external large-scale datasets can also create convergence problem. In Breadcrumbs, data samples are augmented with saved feature vectors from prior epochs and no extra computation.


\section{EMANATE}

In this section, we introduce the data augmentation method that underlies Breadcrumbs.

\subsection{Data Sampling and Decoupling Training}
Consider an image recognition problem with training set ${\cal D} = \{(\mathbf{x}_i, \mathbf{y}_i); i = 1, \ldots , N\}$, where $x_i$ is an example and $y_i \in \{1, \ldots, C\}$ its label, where $C$ is the number of classes. A CNN model combines a feature embedding $\mathbf{z} = f(\mathbf{x}; \theta) \in \mathbb{R}^d$, implemented by several convolutional layers of parameters $\theta$, and a classifier $g(\mathbf{z}) \in [0,1]^C$ that operates on the embedding to produce a class prediction $\hat{y} = \arg \max_i g_i(\mathbf{z})$. Standard (image-balanced) CNN training relies on mini-batch SGD, where each batch is randomly sampled from ${\cal D}$.  A class $j$ of $n_j$ training example has probability $\frac{n_j}{N}$ of being represented in the batch. Without loss of generality, we assume classes sorted by decreasing cardinality, i.e. $n_i \leq n_j$, $\forall i > j$. 

In the long-tail setting, where $n_1 \gg n_C$, the model is not fully trained on classes of large index $j$ (tail classes) and under-fits. This can be avoided with recourse to non-uniform sampling strategies, the most popular of which is class-balanced sampling. This samples each class with probability $\frac{1}{C}$, over-sampling tail classes, and is particularly successful when the training of embedding and classifier are decoupled~\cite{kang2019decoupling}, which is also simple to implement. The embedding is first trained with image-balanced sampling, and different sampling and structures can then be used for the classifier. In this work, we adopt the popular linear classifier $g(\mathbf{z})=\nu(\mathbf{W}\mathbf{x} + \mathbf{b})$, where $\nu$ is the softmax function, and class-balanced sampling. 

\subsection{Augmentation by Feature back-tracking} \label{sec:augmentation}

Class-balanced sampling over-samples classes of few examples. For a class $j$ with $n_j < N/C$ the over-sampling factor is $\rho=\frac{N}{Cn_j}$. In the long-tail setting, $\rho$ is usually larger than $10$. This heavily resamples the few available samples and can lead to over-fitting, impairing generalization for tail classes. While over-fitting can be combated with data augmentation, traditional image-level methods, such as random cropping, horizontal flipping, or color jittering, make little difference in feature space, because the embedding is trained to be invariant to such transformations. Feature-level augmentations have been investigated in the few-shot setting~\cite{wang2018low,Liu_2018_CVPR,hariharan2017low}, but typically require training of additional models, which add complexity and sometimes have convergence problems. Ideally, the augmentation technique should be adversarial, i.e. increase training difficulty, and require little extra computation. One possibility is to rely on adversarial examples~\cite{goodfellow2014explaining}. However, these require optimization at each training iteration and have large computational cost. In our experience, standard adversarial attacks are  also not effective at improving generalization for tail classes, because they are too close to the few available examples.

In this work, we propose a different adversarial feature-level augmentation strategy, based on {\it feature backtracking.\/} The idea is that the embedding $f(\mathbf{x}; \theta)$, obtained after training converges, is simply the final element in the family of embeddings $f(\mathbf{x}; \theta^e)$ learned from epochs $e \in \{1, \ldots, E\}$, where $E$ is the number of training epochs. 
It follows that a particular image $\mathbf{x}_i$ produces a sequence of feature vectors 
\begin{equation}
    {\cal B}_i = \{ {\bf z}_i^e = f(\mathbf{x}_i; \theta^e) | e \in \{1, \ldots, E\} \}
    \label{eq:trail}
\end{equation}
during the optimization. We equate ${\cal B}_i$ to a trail of {\it bread crumbs\/} that can be backtracked, as illustrated in Figure~\ref{fig:breadcrumb}(right). These bread crumbs can be used to perform data augmentation {\it without\/} added computation.
It suffices to store, at epoch $e$ the set of features
\begin{equation}
    {\cal Z}^e = \{{\bf z}_i^e = f(\mathbf{x}_i; \theta^e)| \mathbf{x}_i \in {\cal D}\}
\end{equation}
produced by the embedding learned at the end of the epoch. This is denoted as the training set {\it snapshot\/} at epoch $e$. 

Since the embedding $f(\mathbf{x}; \theta^e)$ changes with $e$, features from different epochs are usually not aligned in feature space. This may lead to bread crumb trails that are ``all over the place," e.g. because the space has been translated or rotated between epochs. Hence, 
when feature vectors collected at different epochs are to be used together, a {\it class alignment} is recommended to simplify the training. On the other hand, this alignment cannot be too strong, so as not to defeat the purpose of data-augmentation. In particular, the alignment operation should not jeopardize the adversarial nature of the latter. A simple operation, which is shown to satisfy this property in the following section, is to align the mean feature vectors synthesized per class during back-tracking. This consists of splitting ${\cal Z}^e$ into a set of class snapshots, where
\begin{equation}
    {\cal Z}^e_y = \{\mathbf{z}_{i}^e \in {\cal Z}^e | y_i = y\}
    \label{eq:zey}
\end{equation}
is the snapshot of class $y$, compute the mean of each class
\begin{equation}
    \Bar{\mathbf{z}}^e_y=\frac{1}{n_j} \sum_{j = 1}^{n_j} {\mathbf{z}}_{y,j}^e
\end{equation}
where ${\mathbf{z}}_{y,j}^e$ is the $j^{th}$ element of ${\cal Z}^e_y$, and apply 
\begin{equation}
    \mathbf{z}_{y,j}^{e' \rightarrow e}= \mathbf{z}_{y,i}^{e'} - \Bar{\mathbf{z}}^{e'}_j + \Bar{\mathbf{z}}^{e}_j,
    \label{eq:align}
\end{equation}
where $\mathbf{z}_{y,j}^{e' \rightarrow e}$ is the alignment, with respect to snapshot $e$, of the $j^{th}$ feature vector ${\mathbf{z}}_{y,j}^e$ of class $y$ from epoch $e'$. This produces a snapshot {\it transferred  from epoch $e'$ to  $e$\/}
\begin{equation}
    {\cal Z}^{e' \rightarrow e}_y = \{\mathbf{z}_{y,j}^{e' \rightarrow e} | \mathbf{z}_{y,j}^{e'} \in {\cal Z}^{e'}_y\}.
    \label{eq:zee'y}
\end{equation}
This snapshot can then be combined with ${\cal Z}^e_y$ to produce an {\it augmented snapshot of class $y$ for epoch $e$\/}
\begin{equation}
    {\cal A}_y^{e} = {\cal Z}^{e}_y \bigcup  {\cal Z}^{e' \rightarrow e}_y.
\end{equation}
This process is denoted {\it fEature augMentAtioN by bAcktracking wiTh alignmEnt\/} (EMANATE), as ${\cal A}_y^{e}$ backtracks the breadcrumb trails that emanate from class $y$ at epoch $e$.

\subsection{Theoretical justification}

In this section, we provide theoretical motivation for EMANATE as an adversarial data augmentation technique. 
Let $\nu(\mathbf{W}^e\mathbf{z}+\mathbf{b}^e)$ be the linear classifier learned at the end of epoch $e$, i.e. from the snapshots ${\cal Z}^e_y = \{\mathbf{z}_{y,i}^e\}$ of~(\ref{eq:zey}). The corresponding cross-entropy loss is
\begin{equation}
    L({\cal Z}^e, \mathbf{W}^e, \mathbf{b}^e) = \sum_y L_y({\cal Z}^e_y, \mathbf{W}^e, \mathbf{b}^e) 
    \label{eq:loss}
\end{equation}
where 
\begin{equation}
    L_y({\cal Z}^e_y, \mathbf{W}^e, \mathbf{b}^e) = -\frac{1}{|{\cal Z}^e_y|} \Sigma_i \log \nu_y(\mathbf{W}^e\mathbf{z}^e_{y,i}+\mathbf{b}^e), \label{eq:closs}
\end{equation}
is the loss of class $y$ and $\nu_y$ the $y^\text{th}$ element of the softmax output. It is assumed that the classifier is optimal for the training data under this loss, i.e. 
\begin{equation}
    L_y({\cal Z}^e_y, \mathbf{W}^e, \mathbf{b}^e) \leq L_y({\cal Z}^e_y, \mathbf{W}, \mathbf{b}), \quad \forall \, y, \mathbf{W}, \mathbf{b}.
    \label{eq:bound}
\end{equation}
A feature augmentation procedure adds new features to ${\cal Z}^e_y$. It is denoted adversarial when the augmented training set is more challenging than the original.

\begin{Def}
 Consider the augmentation ${\cal A}^e_y$ of the training set snapshot ${\cal Z}_y^e$ from epoch $e$ and class $y$. The augmentation is {\it adversarial\/} with respect to class $y$ if 
 \begin{equation}
     L_y({\cal A}^e_y, \mathbf{W}^e, \mathbf{b}^e) > L_y({\cal Z}^e_y, \mathbf{W}^e, \mathbf{b}^e)
 \end{equation}
 where $L_y(.)$ the loss of (\ref{eq:closs}).
\end{Def}
For low-shot classes $y$, class-balanced sampling replicates the features of ${\cal Z}^e_y$, creating the augmented feature set ${\cal A}^e_y = {\cal Z}^e_y \cup {\cal Z}^e_y$. Since, from (\ref{eq:closs}) 
\begin{eqnarray}
    L_y({\cal A}^e_y, \mathbf{W}^e, \mathbf{b}^e) 
    &=& -\frac{1}{2|{\cal Z}^e_y|} 2\Sigma_i \log \nu_y(\mathbf{W}^e\mathbf{z}^e_{y,i}+\mathbf{b}^e), \nonumber \\
    &=& L_y({\cal Z}^e_y, \mathbf{W}^e, \mathbf{b}^e) 
\end{eqnarray}
we obtain the following corollary.
\begin{Cor}
Class-balanced sampling is not an adversarial feature augmentation strategy.
\end{Cor}

We next consider augmentation with EMANATE. The following lemma establishes a lower bound for the increase of the training loss under this augmentation technique.
\begin{Lemma}
    Consider the augmentation of ${\cal Z}_y^e$ with the snapshot transferred from epoch $e' < e$ by EMANATE,  i.e.
     ${\cal A}_y^e = {\cal Z}_y^e \cup   {\cal Z}^{e' \rightarrow e}_y$, where  ${\cal Z}^{e' \rightarrow e}_y$ is as defined in~(\ref{eq:zee'y}). Then 
     \begin{eqnarray}
        L_y({\cal A}^e_y, \mathbf{W}^e, \mathbf{b}^e) - L_y({\cal Z}^e_y, \mathbf{W}^e, \mathbf{b}^e) &\geq & \nonumber\\
         \frac{L_y({\cal Z}^{e'}_y, \mathbf{W}^{e'}, \mathbf{b}^{e'}) - L_y({\cal Z}^e_y, \mathbf{W}^e, \mathbf{b}^e)}{2},
     \end{eqnarray}
     where ($\mathbf{W}^e, \mathbf{b}^e$) is the classifier of~(\ref{eq:bound}).\footnote{Proof is provided in supplementary material.}
\end{Lemma}

\begin{figure}[t!]
\centering{

	\includegraphics[width=0.45\textwidth]{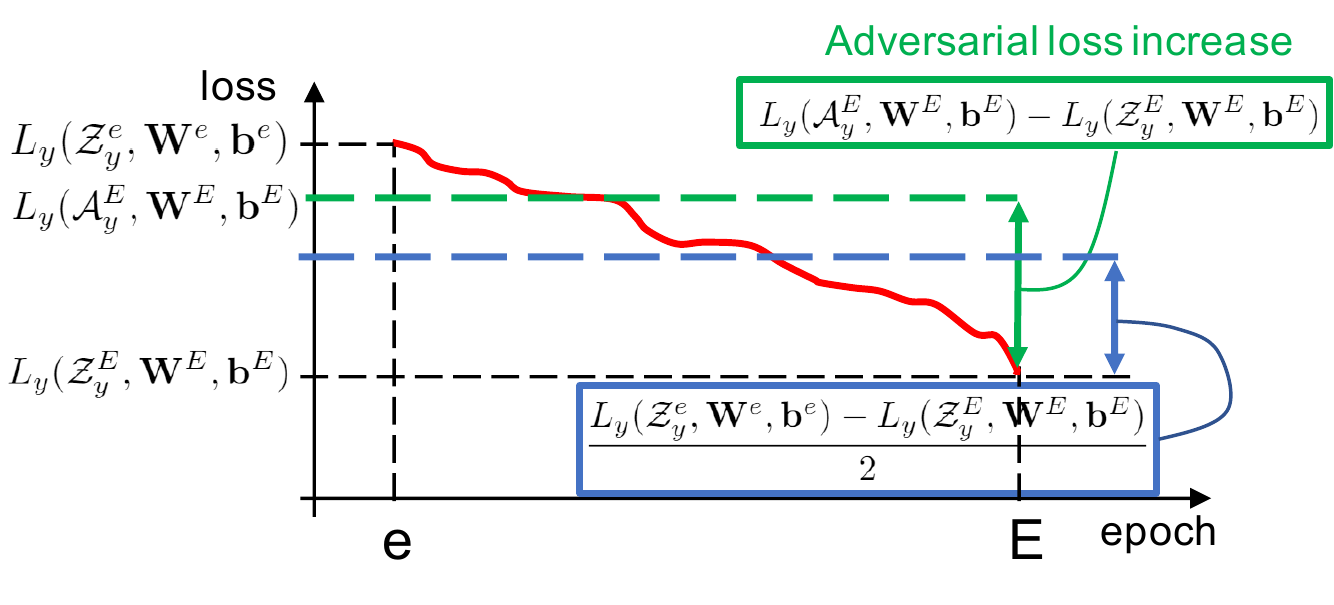}
	}
	\caption{Adversarial nature of EMANATE. The loss increase, between two epochs, due to feature augmentation by EMANATE  is never smaller than half of the training gain (loss decrease) between them.}
	\label{fig:augmetationgain}
\end{figure}

The lemma shows that the adversarial increase of the loss due to the augmentation ($ L_y({\cal A}^e_y, \mathbf{W}^e, \mathbf{b}^e) - L_y({\cal Z}^e_y, \mathbf{W}^e, \mathbf{b}^e)$) is at least half of decrease in the loss of the trained classifier between epochs $e'$ (loss $L({\cal Z}_y^{e'}, \mathbf{W}^{e'}, \mathbf{b}^{e'})$) and $e$ (loss  $L_y({\cal Z}^e_y, \mathbf{W}^e, \mathbf{b}^e$)), i.e. half of what has been gained by training the classifier from epochs $e'$ to $e$. This is illustrated in Figure~\ref{fig:augmetationgain} and leads to the following theorem.

\begin{Them}
EMANATE is an adversarial feature augmentation strategy for any convergent training scheme, i.e. whenever $L_y({\cal Z}^{e'}_y, \mathbf{W}^{e'}, \mathbf{b}^{e'}) > L_y({\cal Z}^e_y, \mathbf{W}^e, \mathbf{b}^e) \forall e' < e$.
\end{Them}
Since successful training requires a convergent training scheme, EMANATE is an adversarial feature augmentation technique for most training procedures of practical interest.

\begin{figure*}[t!]
\centering{
	\includegraphics[width=\textwidth]{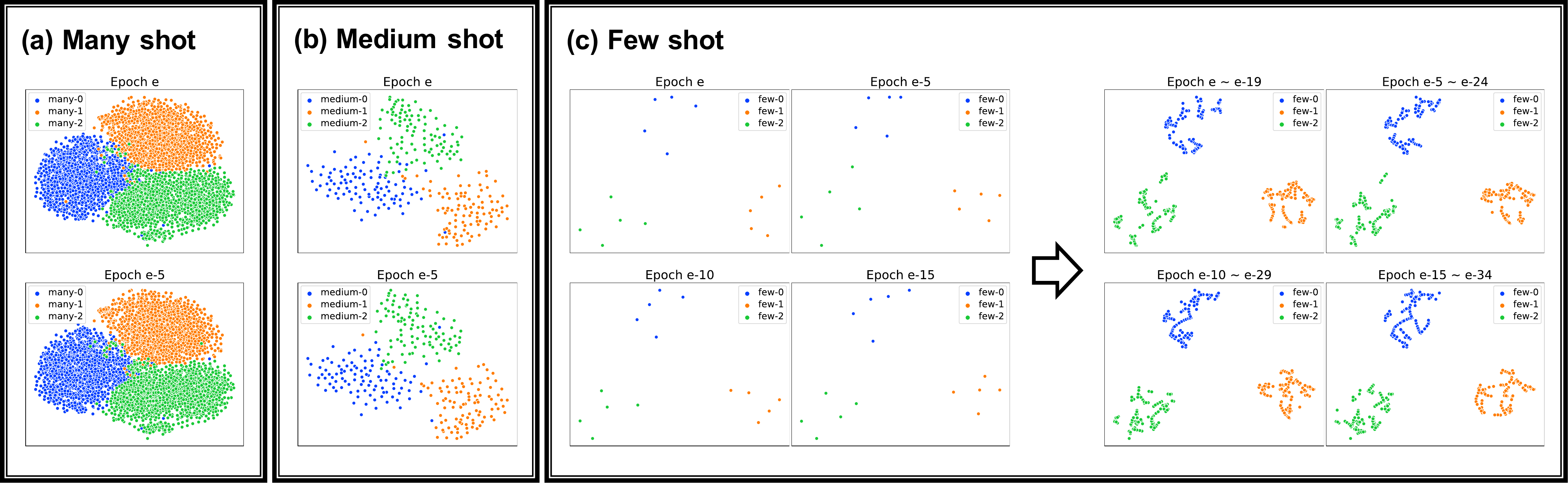}
	}
	\caption{t-SNE visualizations of feature snapshots at different epochs. Many-shot (a) and medium-shot (b) features compose a well-defined geometry that does not change along epochs. Due to the scarcity of data, few-shot features (c, left) fail to hold a consistent geometry along epochs. After augmentation with EMANATE (c, right), the features have more variety and the geometry changes less among epochs.}
	\label{fig:tsne}
\end{figure*}

\subsection{Assembling feature trails}

So far, we have considered the augmentation of ${\cal Z}^{e}_y$ with the transferred snapshot ${\cal Z}^{e' \rightarrow e}_y$.
The augmentation can obviously be repeated for several transferred snapshots $e'$, in order to meet any target number $n_B$ of samples per class at epoch $e$. This is done as follows. Consider a class $j$ with $n_j$ samples.  The features in ${\cal Z}^e_j$ are first selected. If there is a differential to $n_B$, the transferred snapshot ${\cal Z}^{e-1 \rightarrow e}_j$ is selected next. The procedure is repeated until number of the feature vectors reaches $n_B$. If the addition of the final set places the feature cardinality above $n_B$, the necessary number of feature vectors is sampled randomly.
The augmented set of features that emanate from class $j$ at epoch $e$ is then 
\begin{equation}
    {\cal A}^e_j = \bigcup_{k=0}^{K_j-2}{\cal Z}^{e-k \rightarrow e}_j \\
    \bigcup {\tilde{{\cal Z}}}^{e-(K_j-1) \rightarrow e}_j,  
    \label{eq:Fj}
\end{equation}
where $K_j = \left\lceil \frac{n_B}{n_j} \right\rceil,$ and ${\tilde{{\cal Z}}}^{e-K_j-1 \rightarrow e}_j$ is a random sample from ${\cal Z}^{e-K_j-1 \rightarrow e}_j$ of size $n_B - K_j n_j$. The complete training set of epoch $e$ is ${\cal A}^e = \cup_{j=1}^C {\cal A}^e_j$.

The number of snapshots in ${\cal A}^e_j$ depends heavily on the number of examples $n_j$ of the class. 
As shown in Figure~\ref{fig:augmentation_and_hard} (a, right), many-shot classes use a single snapshot, medium-shot classes require snapshots from a few epochs, and few-shot classes require many snapshots to assemble enough training features. However, in all cases, because all feature vectors are already computed during the optimization of the embedding, the only computation required is the mean alignment of~(\ref{eq:align}). This is negligible when compared to the back-propagation computations, making EMANATE nearly computation free, if the necessary snapshots are kept in memory.  In fact, it only necessary to keep in memory the snapshots of classes with $n_j < n_B$. Furthermore, the number $K_j$ of snapshots to be stored adapts to $n_j$, as shown in~(\ref{eq:Fj}). The larger the class, the fewer snapshots are required. In summary, EMANATE has no computational overhead and adapts the memory requirements to the class cardinalities, never requiring more than $n_B$ examples per class. This is the complexity of class-balanced sampling.

Figure~\ref{fig:tsne} shows a t-SNE~\cite{maaten2008visualizing} visualization of training set snapshots collected at different epochs. While the geometry of many- and medium shot classes (Figure~\ref{fig:tsne}(a,b)) is fairly stable across epochs, that of few-shot classes (Figure~\ref{fig:tsne}(c) left) can change significantly, due to data scarcity. EMANATE produces larger clusters with more stable geometry, enabling a more robust training set for the classifier.

\section{Breadcrumbs}

In this section, we investigate two sampling mechanisms based on EMANATE, which are denoted as Breadcrumb sampling. The two mechanisms differ in how the sets  ${\cal A}^e_j$ are collected. In both cases, the two stage training procedure of~\cite{kang2019decoupling} is adopted. In the first stage, the feature extractor $f({\bf x}; \theta)$ and the classifier $\nu({\bf Wx} + b)$ are trained with image balanced sampling. The sets ${\cal A}^e_j, e = \{1, \ldots, E\}$ of class snapshots are collected at each epoch of this stage. In the second stage, the feature extractor $f({\bf x}; \theta)$ is kept fixed and the classifier $\nu({\bf Wx} + b)$ retrained using these sets. As shown in Figure~\ref{fig:breadcrumbsampling}, the two augmentation schemes differ in the classifier update step.



\subsection{\bf Weak Beadcrumb Sampling.}
\begin{figure*}[t!]
\centering{

	\includegraphics[width=1\textwidth]{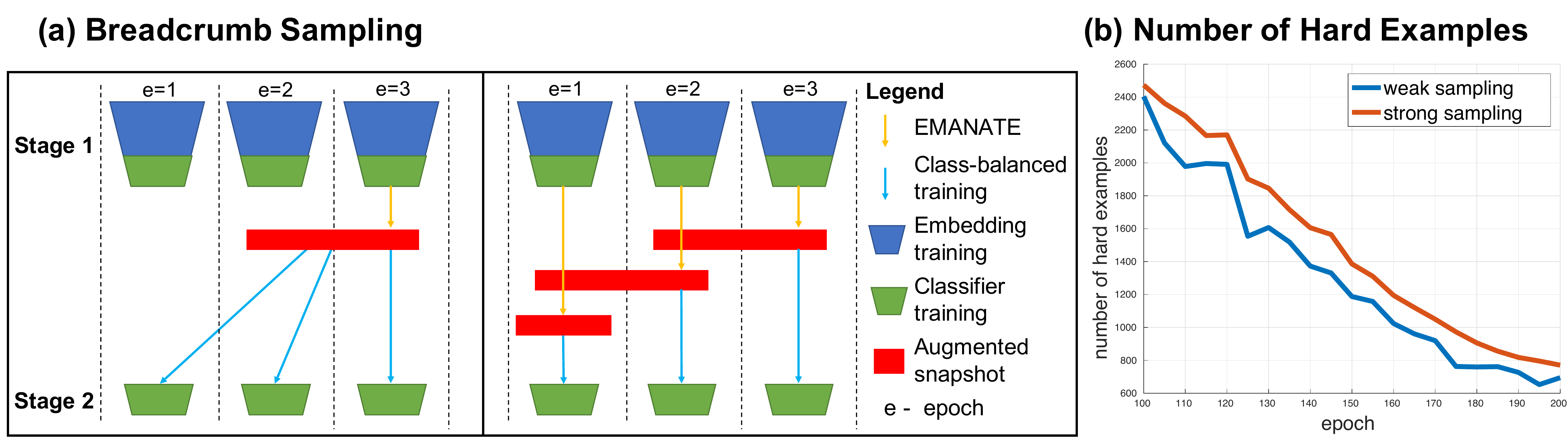}
	}
	\caption{(a) Breadcrumb Sampling relies on EMANATE to collect augmented snapshots ${\cal A}_j^e$ (in red) in a first stage, when the embedding is trained with image-balanced sampling. In a second stage, the classifier is learned with class-balanced training based on these snapshots. In this example $E=3$ and snapshots have length $K_j = 2$ (a single class is shown for simplicity). Left: Weak Breadcrumb Sampling only uses snapshots collected at the end of stage 1. Right: Strong Breadcrumb Sampling uses snapshots collected throughout  stage 1 training. (b) Number of hard examples (loss larger than $5$) in few-shot classes during training, for ResNet-10 on ImageNet-LT. Strong Breadcrumb sampling increases the number of hard examples during training, compared to the weak one. The plot starts at epoch $100$ because early epochs have too many hard examples and dominate the scale.}
	\label{fig:breadcrumbsampling}
\end{figure*}

In the first approach, EMANATE is only applied {\it after convergence\/} of the first stage training. That is,  only the sets ${\cal A}^E_j$ assembled in the {\it final\/} epoch $E$ of the first stage are used to retrain the classifier in the second stage. This is illustrated in the left of Figure~\ref{fig:breadcrumbsampling}, for the case where $E=3$ and augmentation sets span two epochs.
We refer to this sampling technique as {\it Weak Beadcrumb Sampling}, since all snapshots emanate from the feature set produced by the optimal embedding $f(\mathbf{x},\theta^E)$. While this creates some diversity, feature snapshots from neighboring epochs are likely to be similar. This makes the sampling technique less adversarial and therefore ``weak".

\subsection{\bf Strong Beadcrumb Sampling.} \label{sec:evolution}
Strong Beadcrumb Sampling aims to increase feature diversity, so as to create a more adversarial data augmentation. Rather than the augmentation ${\cal A}_j^E$ of the final epoch $E$ of the first stage training, {\it all} augmentations ${\cal A}_j^e, e = \{1, \ldots, E\}$ are saved in that stage, as illustrated in the right of Figure~\ref{fig:breadcrumbsampling}. The classifier retraining of the second stage is then run for $E$ epochs, using the the feature trail sets ${\cal A}_j^e$ collected at epoch $e$ of the first stage to train the classifier in epoch $e$ of the second stage.   
Since each epoch of classifier training contains new data, increasing the difficulty of the classification task, this sampling method is more adversarial and therefore "strong". 
Since the classifier is trained on an evolving feature set ${\cal A}^e$, 

This setting yields a natural selection of the target number $n_B$ of samples per class. To keep the pace of classifier training the same as the embedding training, the size of the dataset should be approximately the same, i.e.
$ n_B = \left\lceil \frac{1}{C}\sum_{j=1}^{C}n_j \right\rceil$. Figure~\ref{fig:augmentation_and_hard}(b), shows that Strong Breadcrumb Sampling increases the number of hard examples in few-shot classes per epoch, when compared to Weak Breadcrumb Sampling. This confirms that it is a more adversarial data augmentation strategy.

\section{Experiments}

In this section, we discuss an evaluation of Breadcrumb sampling.

\subsection{Experimental set-up}
\noindent{\bf Datasets.}
We consider three long-tailed recognition datasets, 
ImageNet-LT~\cite{liu2019large}, Places-LT~\cite{liu2019large} and iNatrualist18~\cite{van2018inaturalist}. ImageNet-LT is a long-tailed version of ImageNet~\cite{imagenet_cvpr09} by sampling a subset following the Pareto distribution with power value $\alpha=6$. 
It contains $115.8$K images from $1000$ categories, with class cardinality ranging from $5$ to $1280$.
Places-LT is a long-tailed version of the Places dataset~\cite{zhou2014learning}. 
It contains $184.5$K images from $365$ categories with class cardinality in $[5, 4980]$.
iNatrualist18 is a long-tailed dataset, which contains $437.5$K images from $8141$ categories with class cardinality in $[2,1000]$.
Following~\cite{liu2019large}, we present classification accuracies for both the entire dataset and three groups of classes: {\it many shot} (more than $100$ training samples), {\it medium shot} (between $20$ and $100$), and {\it few shot} (less than $20$ training samples). 

\noindent{\bf Baselines.}
Following~\cite{liu2019large}, we consider three metric-learning baselines, based on the lifted~\cite{oh2016deep}, focal~\cite{lin2017focal}, and range~\cite{zhang2017range} losses, 
and one state-of-the-art method, FSLwF~\cite{gidaris2018dynamic}, for learning without forgetting.
We also include long-tailed recognition methods designed specifically for 
these datasets,  OLTR~\cite{liu2019large} and Distill~\cite{xiang2020learning}, plus the recent state of the art Decoupling method~\cite{kang2019decoupling}. The model with standard random sampling and end-to-end training is denoted as the {\it Plain Model} for comparison.

\noindent{\bf Training Details.}
ResNet-10 and ResNeXt-50~\cite{he2016deep,xie2017aggregated} are used on ImageNet-LT; ResNet-152 is used on Places-LT; and ResNet-50 is used on iNatrualist18.
The model is trained with SGD, using momentum $0.9$, weight decay $0.0005$, and a learning rate that cosine decays from $0.2$ to $0$. Each iteration uses class-balanced and random sampling mini-batches of size $512$. One epoch is defined when the random sampling iterates over the entire training data. 
Under Strong Breadcrumb Sampling, class-balanced sampling is applied in the initial classifier training epochs, when there are not enough previous epochs to back-track. Codes are attached in supplementary.

\begin{table}
  \caption{Ablation of Breadcrumb components, on the ImageNet-LT. For many-shot $t >100$, for medium-shot $t \in (20,100]$, and for few-shot $t \leq 20$, where $t$ is the number of training samples.}
  \label{tab:ablation}
  \centering
  \scriptsize
  \begin{tabular}{lcccc}
    \toprule
    Method   & Overall & Many-Shot & Medium-Shot & Few-Shot \\
    \midrule
    Decoupling~\cite{kang2019decoupling} & 41.4 & 51.8 & 38.8 & 21.5 \\
    \quad + back-tracking & 41.2 & 50.4 & 38.5 & 23.8 \\
    \quad + class-specific & 41.3 & 50.8 & 38.1 & 24.6 \\
    Weak Breadcrumb & 43.2 & 53.6 & 39.8 & 25.1 \\
    Strong Breadcrumb & {\bf 44.0} & 53.7 & {\bf 41.0} & {\bf 26.4}  \\
    \midrule
    Breadcrumb & {\bf 44.0} & 53.7 & {\bf 41.0} & {\bf 26.4}  \\
    Breadcrumb(var.) & 43.9 & {\bf 53.8} & 40.8 & 26.0  \\
    Breadcrumb(agn.) & 38.5 & 47.3 & 35.6 & 24.0  \\
    \bottomrule
  \end{tabular}
\end{table}

\begin{table*}
  \caption{Results on ImageNet-LT and Places-LT. ResNet-10/152 are used for all methods. For many-shot $t >100$, for medium-shot $t \in (20,100]$, and for few-shot $t \leq 20$, where $t$ is the number of training samples.}
  \label{tab:imagenetlt}
  \centering
  \small
  \begin{tabular}{l|cccc|cccc}
    \toprule
     & \multicolumn{4}{c|}{ImageNet-LT, ResNet-10} & \multicolumn{4}{c}{Places-LT, ResNet-152} \\
    Method   & Overall & Many-Shot & Medium-Shot & Few-Shot & Overall & Many-Shot & Medium-Shot & Few-Shot \\
    \midrule
    Plain Model & 23.5 & 41.1 & 14.9 & 3.6 & 27.2 & {\bf 45.9} & 22.4 & 0.36 \\
    Lifted Loss~\cite{oh2016deep} & 30.8 & 35.8 & 30.4 & 17.9 & 35.2 & 41.1 & 35.4 & 24.0 \\
    Focal Loss~\cite{lin2017focal} & 30.5 & 36.4 & 29.9 & 16.0 & 34.6 & 41.1 & 34.8 & 22.4 \\
    Range Loss~\cite{zhang2017range} & 30.7 & 35.8 & 30.3 & 17.6 & 35.1 & 41.1 & 35.4 & 23.2 \\
    FSLwF~\cite{gidaris2018dynamic} & 28.4 & 40.9 & 22.1 & 15.0 & 34.9 & 43.9 & 29.9 & 29.5 \\
    OLTR~\cite{liu2019large} & 35.6 & 43.2 & 35.1 & 18.5 & 35.9 & 44.7 & 37.0 & 25.3 \\
    Distill~\cite{xiang2020learning} & 38.8 & 47.0 & 37.9 & 19.2 & 36.2 & 39.3 & 39.6 & 24.2 \\
    Decoupling(cRT)~\cite{kang2019decoupling} & 41.4 & 51.8 & 38.8 & 21.5 & 37.9 & 37.8 & 40.7 & 31.8 \\
    \midrule
    Breadcrumb & {\bf 44.0} & {\bf 53.7} & {\bf 41.0} & {\bf 26.4} & {\bf 39.3} & 40.6 & {\bf 41.0} & {\bf 33.4} \\
    \bottomrule
  \end{tabular}
\end{table*}

\begin{table*}
\parbox{.6\textwidth}{
  \caption{Results on ImageNet-LT, ResNeXt-50. For many-shot $t >100$, for medium-shot $t \in (20,100]$, and for few-shot $t \leq 20$, where $t$ is the number of training samples.}
  \label{tab:imagenetlt2}
  \centering
  \footnotesize
  \begin{tabular}{l|cccc}
    \toprule
    Method   & Overall & Many-Shot & Medium-Shot & Few-Shot \\
    \midrule
    OLTR~\cite{liu2019large} & 41.9 & 51.0 & 40.8 & 20.8 \\
    Decoupling(NCM)~\cite{kang2019decoupling} & 47.3 & 56.6 & 45.3 & 28.1  \\
    Decoupling(cRT)~\cite{kang2019decoupling} & 49.6 & 61.8 & 46.2 & 27.4  \\
    Decoupling($\tau$)~\cite{kang2019decoupling} & 49.4 & 59.1 & 46.9 & 30.7  \\
    Decoupling(LWS)~\cite{kang2019decoupling} & 49.9 & 60.2 & 47.2 & 30.3 \\
    \midrule
    Breadcrumb & {\bf 51.0} & {\bf 62.9} & {\bf 47.2} & {\bf 30.9} \\
    \bottomrule
  \end{tabular}
  }
  \hfill
\parbox{.35\textwidth}{
  \caption{Results on the iNaturalist 2018. All methods are implemented with ResNet-50.}
  \small
  \centering
  \label{tab:inat}
  \begin{tabular}{lc}
    \toprule
    Method   & Accuracy \\
    \midrule
    CB-Focal~\cite{cui2019class} & 61.1 \\
    LDAM+DRW~\cite{cao2019learning} & 68.0 \\
    Decoupling(cRT)~\cite{kang2019decoupling} & 68.2 \\
    Decoupling($\tau$)~\cite{kang2019decoupling} & 69.3 \\
    Decoupling(LWS)~\cite{kang2019decoupling} & 69.5 \\
    \midrule
    Breadcrumb & {\bf 70.3} \\
    \bottomrule
  \end{tabular}
}
\end{table*}



\subsection{Ablation Study}
Several ablations were performed to study the effectiveness of the various components of Breadcrumb. In this study, all models are trained and evaluated on the training and test set of ImageNet-LT, respectively, using a ResNet-10 backbone.

\noindent{\bf Component ablation.}
Starting from the baseline Decoupling (cRT)~\cite{kang2019decoupling} method, we incrementally add feature back-tracking, class-specific augmentation, class alignment (leading to Weak Breadcrumb Sampling), and Strong Breadcrumb Sampling. Results are shown in Table~\ref{tab:ablation}. When only back-tracking is applied, all snapshots are collected from the last $10$ epochs of image-balanced training (first stage), and the classifier trained (in the second stage) using this feature set and class-balanced sampling. No class alignment is applied. Compared to the baseline, back-tracking gives a reasonable gain on few-shot classes but harms many-shot performance. This can be explained by the fact that, for many-shot classes, features from the final epoch are replaced by those from prior epochs. Since the corresponding embeddings are sub-optimal, the augmented features are inferior to the final ones. This, however, is not the case in few-shot, where augmented features replace {\it duplicated\/} features. 

The combination of back-tracking and class-specific augmentation, where different classes have different back-tracking lengths, is denoted as ``+ class-specific'' in Table~\ref{tab:ablation}. Surprisingly, without class alignment, the performance on many-shot does not improve, even though no augmented features are introduced into those classes. We believe this is due to the fact that when few-shot features are augmented without alignment, those augmented features take up position in feature space that should not be assigned to them. This decreases the accuracy of many-shot classes. 
When class-alignment is applied (Weak Breadcrumb Sampling) we observe an improvement over all class partitions, with gains of $1.8\%$ (Many), $1\%$ (Medium), and $3,6\%$ (Few-Shot) and an overall improvement of $1.8\%$ over the baseline. Finally, Strong Breadcrumb Sampling enables another $0.8\%$ overall gain, for a total gain of $2.6\%$ over the baseline.

\noindent{\bf Class alignment ablation.}
Since alignment makes a significant difference, we considered three different alignment choices. In Sec~\ref{sec:augmentation}, only class-specific mean alignment is presented. It is also possible to align the feature variances. This is denoted as Breadcrumb(var.) in Table~\ref{tab:ablation} and has a negligible difference. Hence, we only apply mean alignment unless otherwise noted.
Another possibility is class-agnostic alignment, where only one mean is computed over all classes. This is listed as Breadcrumb(agn.) in Table~\ref{tab:ablation}. Its poor performance implies that class-agnostic alignment cannot fully eliminate the differences between epochs.

\subsection{Comparison to the state of the art}
Table~\ref{tab:imagenetlt} presents a final comparison to the methods in the literature on ImageNet-LT, using a ResNet-10, and Places-LT, using a ResNet-152. In these experiments we use Strong Breadcrumb Sampling, which is shown to outperform all other methods on both datasets. It achieves  the  best  performance  on $5$ of the $6$ partitions and is always better than the next overall best performer (Decoupling(cRT)). It is only outperformed by the Plain Model on the Many-Shot split of Places-LT, where this model severely overfits to the Many-Shot classes, basically ignoring the Few-Shot ones, and achieving overall performance $12.1\%$ weaker than Breadcrumb. Compared to the best models Breadcrumb also achieves significant gains on few-shot classes, especially on ImageNet-LT, where it beats the next best method by $4.9\%$. This suggests that previous methods over-fit for few-shot classes, a problem that is mitigated by the introduction of EMANATE and Strong Breadcrumb Sampling. 
Table~\ref{tab:imagenetlt2} shows that these results are fairly insensitive to the backbone network. Breadcrumb achieves the best overall performance and the best performance on all partitions with a ResNeXt-50 backbone. Finally Table~\ref{tab:inat} shows that Breadcrumb again achieves the overall best results for a ResNet-50 on iNaturalist.

\section{Conclusion}
This work discussed the long-tailed recognition problem. A new augmentation framework, Breadcrumb, was proposed to increase feature variety and  classifier robustness. Breadcrumb is based on EMANATE, a feature back-tracking procedure that aligns features vectors produced across several epochs of embedding training,  to compose a class-balanced feature set for training the classifier at the top of the network. It is inspired by the the recent success of class-balanced training schemes. However, unlike previous schemes, it is shown to be an adversarial sampling scheme, a property that encourages better generalization. A comparison of two sampling schemes based on EMANATE confirmed this property, resulting in best performance for the Strong Breadcrumb Sampling technique, where feature snapshots are collected while the embedding is evolving. Breadcrumb was shown to achieve state-of-the-art performance on three popular long-tailed datasets with different CNN backbones. Furthermore, Breadcrumb introduces no extra model, which means that it adds no computational overhead or convergence issues to the baseline model.



{\small
\bibliographystyle{ieee_fullname}
\bibliography{bread}
}

\end{document}